\documentclass[10pt,twocolumn,letterpaper]{article}

\usepackage{iccv}
\usepackage{times}
\usepackage{epsfig}
\usepackage{graphicx}
\usepackage{amsmath}
\usepackage{amssymb}

\newtheorem{theorem}{Theorem}[section]

\newtheorem{lemma}[theorem]{Lemma}

\usepackage[breaklinks=true,bookmarks=false]{hyperref}
\usepackage{authblk}

\usepackage[breaklinks=true,bookmarks=false]{hyperref}

\iccvfinalcopy 

\def\iccvPaperID{464} 

\ificcvfinal\fi

\begin{document}

\title{A Closed-form Solution to Universal Style Transfer}


\author[1]{Ming Lu \thanks{This work was done when Ming Lu was an intern at Intel Labs China, supervised by Anbang Yao who is responsible for correspondence.}}
\author[1]{Hao Zhao}
\author[2]{Anbang Yao}
\author[2]{Yurong Chen}
\author[3]{Feng Xu}
\author[1]{Li Zhang}

\affil[1]{Department of Electronic Engineering, Tsinghua University}
\affil[2]{Intel Labs China}
\affil[3]{BNRist and School of Software, Tsinghua University}

\affil[ ]{\small {\{lu-m13@mails,zhao-h13@mails,feng-xu@mail,chinazhangli@mail\}.tsinghua.edu.cn}}
\affil[ ]{\small {\{anbang.yao, yurong.chen\}@intel.com}}

\maketitle
\ificcvfinal\thispagestyle{empty}\fi

\begin{abstract}
	Universal style transfer tries to explicitly minimize the losses in feature space, thus it does not require training on any pre-defined styles. It usually uses different layers of VGG network as the encoders and trains several decoders to invert the features into images. Therefore, the effect of style transfer is achieved by feature transform. Although plenty of methods have been proposed, a theoretical analysis of feature transform is still missing. In this paper, we first propose a novel interpretation by treating it as the {\bf optimal transport} problem. Then, we demonstrate the relations of our formulation with former works like Adaptive Instance Normalization (AdaIN) and Whitening and Coloring Transform (WCT). Finally, we derive a closed-form solution named {\bf Optimal Style Transfer} (OST) under our formulation by additionally considering the content loss of Gatys. Comparatively, our solution can preserve better structure and achieve visually pleasing results. It is simple yet effective and we demonstrate its advantages both quantitatively and qualitatively. Besides, we hope our theoretical analysis can inspire future works in neural style transfer. Code is available at \url{https://github.com/lu-m13/OptimalStyleTransfer}.
\end{abstract}

\section{Introduction}

A variety of methods on neural style transfer have been proposed since the seminal work of Gatys \cite{gatys2016image}. These methods can be roughly categorized into image optimization and model optimization \cite{jing2017neural}. Methods based on image optimization directly obtain the stylized output by minimizing the content loss and style loss. The style loss can be defined by Gram matrix \cite{gatys2016image}, histogram \cite{risser2017stable}, or Markov Random Fields (MRFs) \cite{li2016combining}. Contrary to that, methods based on model optimization try to train neural networks on large datasets like COCO \cite{lin2014microsoft}. The training loss can be defined as perceptual loss \cite{johnson2016perceptual} or MRFs loss \cite{li2016precomputed}. Subsequent works \cite{chen2017stylebank,dumoulin2017learned,zhang2018multi} further study the problem of training one network for multiple styles. Recently, \cite{huang2017arbitrary} proposes to use AdaIN as feature transform to train one network for arbitrary styles. Apart from image and model optimization, many other works study the problems of semantic style transfer \cite{lu2017decoder,liao2017visual,champandard2016semantic}, video style transfer \cite{huang2017real,chen2017coherent,ruder2016artistic,ruder2018artistic}, portrait style transfer \cite{selim2016painting}, and stereoscopic style transfer \cite{chen2018stereoscopic}. \cite{jing2017neural} provides a thorough review of the works on style transfer.

In this paper, we study the problem of universal style transfer \cite{li2017universal}. Our motivation is to explicitly minimize the losses defined by Gatys \cite{gatys2016image}. Therefore, our approach does not require training on any pre-defined styles. Similar to WCT \cite{li2017universal}, our method is also based on a multi-scale encoder-feature transform-decoder framework. We use different layers of VGG network \cite{simonyan2014very} as the encoders and train the decoders to invert features into images. The effect of style transfer is achieved by feature transform between encoder and decoder. Therefore, the key to universal style transfer is feature transform. In this work, we focus on the theoretical analysis of feature transform and propose a new closed-form solution.

Although AdaIN \cite{huang2017arbitrary} trains its decoder on a large dataset of style images, AdaIN itself is also a feature transform method. It considers the feature of each channel as a Gaussian distribution and assumes the channels are independent. For each channel, AdaIN first normalizes the content feature and then matches it to the style feature. This means it only matches the diagonal elements of the covariance matrices. WCT \cite{li2017universal} proposes to use whitening and coloring as feature transform. Compared with AdaIN, WCT improves the results by matching all the elements of covariance matrices. Since the channels of deep Convolutional Neural Networks (CNNs) are correlated, the non-diagonal elements are essential to represent the style. However, WCT only matches the covariance matrices, which shares similar spirits with minimizing the style loss of Gatys. It does not consider the content loss and cannot well preserve the image structure. Moreover, multiplying an orthogonal matrix between the whitening and coloring matrices can also match the covariance matrices, which has been pointed out by \cite{li2018learning}.

\cite{li2017demystifying} shows that matching Gram matrices is equivalent to minimizing the Maximum Mean Discrepancy (MMD) with the second order polynomial kernel. However, it does not give a closed-form solution. Instead, our work reformulates style transfer as an optimal transport problem. Optimal transport tries to find a transformation that matches two high-dimensional distributions. For neural style transfer, considering the neural feature in each activation as a high dimension sample, we assume the samples of content and style images are from two Multivariate Gaussian (MVG) distributions. Style transfer is equivalent to transforming the content samples to fit the distribution of style samples. Assuming the transformation is linear, we find that both AdaIN and WCT are special cases of our formulation. Although \cite{li2018learning} also assumes the transformation is linear, it still follows the whitening and coloring pipeline and trains two meta networks for whitening and coloring matrices. Contrary to that, we directly find the transformation under the optimal transport formulation.

As we have described above, there are still infinite transformations, for example, multiplying an orthogonal matrix between the whitening and coloring matrices can also be the solution. Therefore, we seek for a transformation, which additionally minimizes the difference between transformed feature and original content feature. This shares similar spirits with minimizing the content loss of Gatys \cite{gatys2016image}. We prove that a unique closed-form solution named Optimal Style Transfer (OST) can be found, once considering the content loss. We show the detailed proof of OST in the method part. Since OST further considers the content loss, it can preserve better structures compared with WCT.

Our contributions can be concluded as follows:

1. We present a novel interpretation of neural style transfer by treating it as an optimal transport problem and elucidate the theoretical relations of our interpretation with former works on feature transform, for example, AdaIN and WCT.

2. We find the unique closed-form solution named OST under the optimal transport interpretation by additionally considering the content loss.

3. Our closed-form solution preserves better structures and achieves visually pleasing results.

\begin{figure*}
	\begin{center}
		\includegraphics[width=0.8\linewidth]{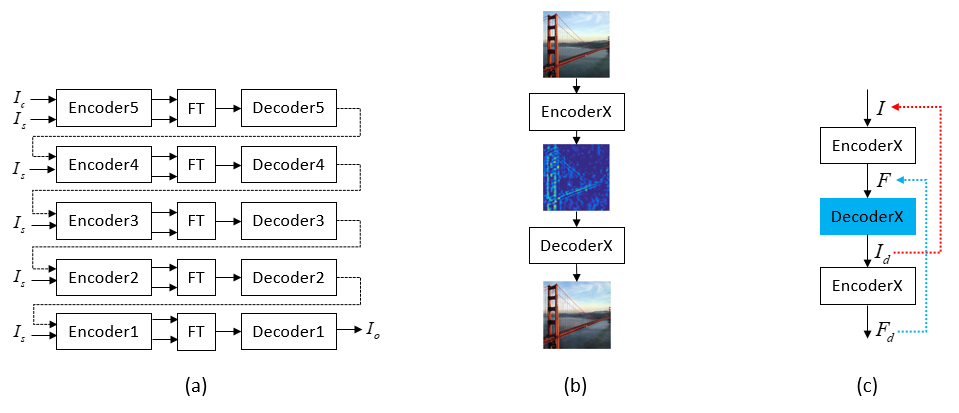}
	\end{center}
	\vspace{-3mm}
	\caption{(a) The pipeline of OST for universal style transfer. First, we extract features using the encoder for content image and style image. Then we use the feature transform method to obtain the stylized feature. Finally, the decoder inverts the stylized feature into image. The output of top layer is used as the input content image for the bottom layer. (b) The decoder inverts the feature of a certain layer to the image. Although \cite{gu2018arbitrary,sheng2018avatar} propose to train the decoder to invert the feature to its bottom layer's feature, which might be more efficient, we use the image decoder in this work since decoder is not our contribution. (c) We use the feature loss (denoted by the blue arrow) and the reconstruction loss (denoted by the red arrow) to train the DecoderX (X=1,2,...,5).}
	\label{fig:pipeline}
\end{figure*} 

\section{Related Work}

{\bf Image Optimization.} Methods based on image optimization directly obtain the stylized output by minimizing the content loss and style loss defined in the feature space. The optimization is usually based on back-propagation. \cite{gatys2015texture,gatys2016image} propose to use Gram matrix to define the style of an example image. \cite{li2016combining} improves the results by combining MRFs with Convolutional Neural Networks. \cite{champandard2016semantic} uses the semantic masks to define the style losses within corresponding regions. In order to improve the results for portrait style transfer, \cite{selim2016painting} proposes to modify the feature maps to transfer the local color distributions of the example painting onto the content image. This is similar to the gain map proposed by \cite{shih2014style}. \cite{gatys2017controlling} studies the problem of controlling the perceptual factors during style transfer. \cite{risser2017stable} improves the results of neural style transfer by incorporating the histogram loss. \cite{ruder2016artistic} incorporates the temporal consistency loss into the optimization for video style transfer. Since all the above methods solve the optimization by back-propagation, they are intrinsically time-consuming.

{\bf Model Optimization.} In order to solve the speed bottleneck of back-propagation, \cite{johnson2016perceptual,li2016precomputed} propose to train a feed-forward network to approximate the optimization process. Instead of optimizing the image, they optimize the parameters of the network. Since it is tedious to train one network for each style, \cite{chen2017stylebank,dumoulin2017learned,zhang2018multi} further study the problem of training one network for multiple styles. Later, \cite{chen2016fast} presents a method based on patch swap for arbitrary style transfer. First, the content and style images are forwarded through the deep neural network to extract features. Then the style transfer is formulated as neural patch swap to get the reconstructed feature map. This feature map is inverted by the decoder network to image space. Since then, the framework of encoder-feature transform-decoder has been widely explored for arbitrary style transfer. \cite{huang2017arbitrary} uses AdaIN as the feature transform and trains the decoder over large collections of content and style images. \cite{li2018learning} trains two meta networks for the whitening and coloring matrices, following the formulation of WCT \cite{li2017universal}. Many other works also extend neural style transfer to video \cite{chen2017coherent,huang2017real,ruder2018artistic} and stereoscopic style transfer \cite{chen2018stereoscopic}. These works usually jointly train additional networks apart from the style transfer network.

{\bf Universal Style Transfer.} Universal style transfer \cite{li2017universal} is also based on the framework of encoder-feature transform-decoder. Unlike AdaIN \cite{huang2017arbitrary}, it does not require network training on any style image. It directly uses different layers of VGG network as the encoders and train the decoders to invert the feature into image. The style transfer effect is achieved by feature transform. \cite{chen2016fast} replaces the patches of content feature by the most similar patches of style feature. However, the nearest neighbor search achieves less transfer effect since it tends to preserve the original appearance. AdaIN considers the activation of each channel as a Gaussian distribution and matches the content and style images through mean and variance. However, since the channels of CNN are correlated, AdaIN cannot achieve visually pleasing transfer effect. WCT \cite{li2017universal} proposes to use feature whitening and coloring to match the covariance matrices of style and content images. However, as pointed out by \cite{li2018learning}, WCT is not the only approach to matching the covariance matrices. \cite{sheng2018avatar} proposes a method to combine patch match with WCT and AdaIN. Instead of finding the nearest neighbor by the original feature, \cite{sheng2018avatar} conducts it using the projected feature. These projected feature can be generated by AdaIN or WCT. However, above methods all fail to give a theoretical analysis of feature transform. The key observation of current works like WCT is matching the covariance matrices, which is not enough to find a good solution.

\section{Motivation}
\label{sec:moti}

The pipeline of OST is shown in Figure \ref{fig:pipeline}. It is similar to WCT \cite{li2017universal}. We use different layers of the pre-trained VGG network as the encoders. For every encoder, we train the corresponding decoder to invert the feature into image as illustrated by Figure \ref{fig:pipeline} (b, c). Although \cite{gu2018arbitrary,sheng2018avatar} propose to train the decoder to invert the feature to its bottom layer's feature, which might be more efficient, we use the image decoder \cite{li2017universal} in this work since the framework is not our contribution.

We start to study the problem of feature transform by reformulating neural style transfer as the optimal transport problem. We denote the content image as ${I_c}$ and the style image as ${I_s}$. For the features of content image and style image, we denote them as ${F_c} \in {R^{C \times {H_c}{W_c}}}$ and ${F_s} \in {R^{C \times {H_s}{W_s}}}$ separately, where ${{H_c}{W_c}}$ and ${{H_s}{W_s}}$ are the numbers of activations and $C$ is the number of channels. We view the columns of ${F_c}$ and ${F_s}$ as samples from two Multivariate Gaussian (MVG) distributions $N({\mu _c},{\Sigma _c})$ and $N({\mu _s},{\Sigma _s})$, where ${\mu _c},{\mu _s} \in {R^C}$ are the mean vectors and ${\Sigma _c},{\Sigma _s} \in {R^{C \times C}}$ are the variance matrices. We further denote the sample from content distribution as $u$ and the sample from style distribution as $v$. Therefore, $u \sim N({\mu _c},{\Sigma _c})$ and $v \sim N({\mu _s},{\Sigma _s})$. Assuming the optimal transformation is linear, we can represent it as follows.

\begin{equation}
t(u) = T(u - {\mu _c}) + {\mu _s}
\label{eq:1}
\end{equation}

Where $T \in {R^{C \times C}}$ is the transformation matrix. Since we assume the features are from two MVG distributions, $T$ must meet the following equation to match two MVG distributions.

\begin{equation}
T{\Sigma _c}{T^T} = {\Sigma _s}
\label{eq:2}
\end{equation}

\begin{figure}[t]
	\begin{center}
		\includegraphics[width=1.0\linewidth]{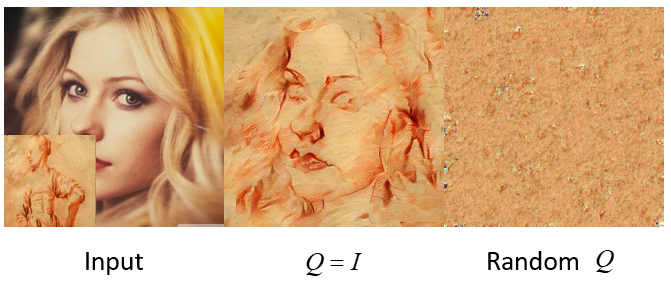}
	\end{center}
	\vspace{-3mm}
	\caption{Style transfer results of $T = \Sigma _s^{1/2}Q\Sigma _c^{ - 1/2}$, where $Q$ is a unite orthogonal matrix. Although $T$ satisfies Eq. \ref{eq:2}, the results vary significantly.}
	\label{fig:randomQ}
\end{figure}

Where ${T^T}$ is the transpose of $T$. When Eq. \ref{eq:2} is satisfied, we can obtain $t(u) \sim N({\mu _s},{\Sigma _s})$. We then demonstrate the relations of our formulation with AdaIN \cite{huang2017arbitrary} and WCT \cite{li2017universal}. We denote the diagonal matrices of ${\Sigma _c}$ and ${\Sigma _s}$ as ${D_c}$ and ${D_s}$ separately. {\bf For AdaIn}, the transformation matrix $T = {D_s}./{D_c}$, where $./$ denotes the element-wise division. Therefore, AdaIN does not satisfy Eq. \ref{eq:2} since it ignores the correlation of channels. Only the diagonal elements are matched by AdaIN. {\bf As for WCT}, we can find that the transformation matrix $T = \Sigma _s^{1/2}\Sigma _c^{ - 1/2}$. Since both $\Sigma _s^{1/2}$ and $\Sigma _c^{ - 1/2}$ are symmetric matrices, WCT satisfies Eq. \ref{eq:2}. However, WCT is not the only solution to Eq. \ref{eq:2} because $T = \Sigma _s^{1/2}Q\Sigma _c^{ - 1/2}$, where $Q$ is a unite orthogonal matrix, is a family of solutions to Eq. \ref{eq:2}. This has also been pointed out by \cite{li2018learning}. Theoretically, there are infinite solutions, considering only Eq. \ref{eq:2}. We show the style transfer results of multiplying a random unite orthogonal matrix to the whitening matrix in Figure \ref{fig:randomQ}. As can been seen, although $T = \Sigma _s^{1/2}Q\Sigma _c^{ - 1/2}$ satisfies Eq. \ref{eq:2}, the style transfer results vary significantly.

Our motivation is to find an optimal solution by additionally considering the content loss of Gatys. Therefore, our formulation can be represented as follows, where $E$ represents the expectation.

\begin{equation}
\begin{aligned}
T = \mathop {\arg \min }\limits_T E(||t(u) - u||_2^2) \\
s.t. \: \: \: \: t(u) = T(u - {\mu _c}) + {\mu _s} \\
\: \: \: \: T{\Sigma _c}{T^T} = {\Sigma _s} 
\label{eq:3}
\end{aligned}
\end{equation}

\section{Method}

In this part, we derive the closed-form solution to Eq. \ref{eq:3}. We substitute Eq. \ref{eq:1} to the expectation term of Eq. \ref{eq:3} and obtain:

\begin{equation}
E[{(T(u - {\mu _c}) + {\mu _s} - u)^T}(T(u - {\mu _c}) + {\mu _s} - u)]
\label{eq:4}
\end{equation}

We denote ${u^ * } = u - {\mu _c}$, ${v^ * } = T{u^ * }$ and $\delta  = {\mu _s} - {\mu _c}$. Therefore, we can get ${u^ * } \sim N(0,{\Sigma _c})$ and ${v^ * } \sim N(0,{\Sigma _s})$. Besides, $\delta$ is a constant C-dimensional vector. Using ${u^ * }$, ${v^ * }$ and $\delta$, we can re-write Eq. \ref{eq:4} as:

\begin{equation}
E[{({v^ * } + \delta  - {u^ * })^T}({v^ * } + \delta  - {u^ * })]
\label{eq:5}
\end{equation}

We further expand Eq. \ref{eq:5} to:

\begin{equation}
\begin{aligned}
E[{v^{ * T}}{v^ * } + {\delta ^T}{v^ * } - {u^ * }^T{v^ * } + {v^{ * T}}\delta  + {\delta ^T}\delta  \\
- {u^ * }^T\delta  - {v^{ * T}}{u^ * } - {\delta ^T}{u^ * } + {u^ * }^T{u^ * }]
\label{eq:6}
\end{aligned}
\end{equation}

Since ${u^ * } \sim N(0,{\Sigma _c})$, ${v^ * } \sim N(0,{\Sigma _s})$ and $\delta$ is a constant C-dimensional vector, we can get $E[{\delta ^T}{v^ * }] = E[{v^{ * T}}\delta ] = 0$ and $E[{u^ * }^T\delta ] = E[{\delta ^T}{u^ * }] = 0$. Besides, $E[{\delta ^T}\delta ]$ is also constant. Therefore, minimizing Eq. \ref{eq:6} is equivalent to minimizing Eq. \ref{eq:7}:

\begin{equation}
\begin{aligned}
E[{v^{ * T}}{v^ * } + {u^ * }^T{u^ * } - {u^ * }^T{v^ * } - {v^{ * T}}{u^ * }]
\label{eq:7}
\end{aligned}
\end{equation}

Using the representation of matrix trace, Eq. \ref{eq:7} can be rewritten as follows. 

\begin{equation}
\begin{aligned}
tr(E[{v^ * }{v^{ * T}} + {u^ * }{u^ * }^T - {v^ * }{u^ * }^T - {u^ * }{v^{ * T}}])
\label{eq:8}
\end{aligned}
\end{equation}

Where $tr$ means the trace of a matrix. Since $E[{v^ * }{v^{ * T}}] = {\Sigma _s}$, $E[{u^ * }{u^ * }^T] = {\Sigma _c}$ and $E[{v^ * }{u^ * }^T] = E[{u^ * }{v^{ * T}}] = \phi $, where $\phi$ denotes the covariance matrix of $v^ *$ and $u^ *$, the solution to Eq. \ref{eq:3} can be reformulated as follows.

\begin{equation}
\begin{aligned}
T = \mathop {\arg \max }\limits_T (tr(\phi ))
\label{eq:9}
\end{aligned}
\end{equation}

Next, we introduce a lemma, which has been proved by \cite{olkin1982distance}. We do not repeat the proof due to limited space. The lemma can be concluded as follows.

\begin{lemma}
	Given two high-dimensional distributions $X$ and $Y$, where $X \sim N(0,{\Sigma _{11}})$ and $Y \sim N(0,{\Sigma _{22}})$, we define the distribution of $(X,Y)$ as $N(0,\Sigma )$, where $\Sigma $ can be represented as follows.
	
	\begin{equation}
	\begin{aligned}
	\Sigma  = \left( {\begin{array}{*{20}{c}}
		{{\Sigma _{11}}}&\phi \\
		{{\phi ^T}}&{{\Sigma _{22}}}
		\end{array}} \right)
	\label{eq:10}
	\end{aligned}
	\end{equation}
	
	The problem of $\max (tr(2\phi ))$ has a unique solution, which can be represented as:
	
	\begin{equation}
	\begin{aligned}
	\phi  = {\Sigma _{11}}\Sigma _{22}^{1/2}{(\Sigma _{22}^{1/2}{\Sigma _{11}}\Sigma _{22}^{1/2})^{ - 1/2}}\Sigma _{22}^{1/2}
	\label{eq:11}
	\end{aligned}
	\end{equation}
	
\end{lemma}

With the above lemma, let $X = {v^ * }$, $Y = {u^ * }$, ${\Sigma _{11}} = {\Sigma _s}$ and ${\Sigma _{22}} = {\Sigma _c}$, we can obtain the solution to Eq. \ref{eq:9}, which can be represented as $\phi  = {\Sigma _s}\Sigma _c^{1/2}{(\Sigma _c^{1/2}{\Sigma _s}\Sigma _c^{1/2})^{ - 1/2}}\Sigma _c^{1/2}$. We rewrite the covariance matrix as $\phi  = E[{v^ * }{u^{ * T}}] = E[{v^ * }{({T^{ - 1}}{v^ * })^T}] = E[{v^ * }{v^{ * T}}]{({T^{ - 1}})^T} = {\Sigma _s}{({T^{ - 1}})^T}$. Therefore, we can get ${({T^{ - 1}})^T} = \Sigma _c^{1/2}{(\Sigma _c^{1/2}{\Sigma _s}\Sigma _c^{1/2})^{ - 1/2}}\Sigma _c^{1/2}$. Then the final $T$ can be represented as Eq. \ref{eq:12}.

\begin{equation}
\begin{aligned}
T = \Sigma _c^{ - 1/2}(\Sigma _c^{1/2}{\Sigma _s}\Sigma _c^{1/2})^{1/2}\Sigma _c^{ - 1/2}
\label{eq:12}
\end{aligned}
\end{equation}  

{\bf Remarks:} The final solution of our method is very simple. Since our method additionally considers the content loss, we can preserve better structure compared with WCT. Contrary to former works, we provide a complete theoretical proof of the proposed method. The relations of our method with former works are also demonstrated. We believe both the closed-form solution and the theoretical proof will inspire future works in neural style transfer. 

\section{Results}

In this section, we first qualitatively compare our method with Gatys \cite{gatys2016image}, Patch Swap \cite{chen2016fast}, AdaIN (with our decoder) \cite{huang2017arbitrary}, AdaIN+ (with their decoder) \cite{huang2017arbitrary}, and WCT \cite{li2017universal} in Section \ref{sec:quali}. Then we provide a quantitative comparison of our method against Gatys, Patch Swap, AdaIN, AdaIN+ and WCT in Section \ref{sec:quanti}. Following former works, we also show results of linear interpolation and semantic style transfer in Section \ref{sec:more}. Finally, we discuss the limitations of our method in Section \ref{sec:limit}.

{\bf Parameters:} We train the decoders on the COCO dataset \cite{lin2014microsoft}. The weight to balance the feature loss and reconstruction loss in Eq. \ref{eq:13} is set to 1 as \cite{li2017universal}. For the results in this work, the resolution of the input is fixed as $512 \times 512$.

{\bf Performance:} We implement the proposed method on a server with an NVIDIA Titan Xp graphics card. The processing speed comparison is listed in Table \ref{tab:perf} under the input resolution of $512 \times 512$. We do the comparison with the published implementations on our server, which might result in slight differences with the papers.

\begin{table}
	\begin{tabular}{|c|c|c|c|c|c|}
		\hline
		Gatys   & Patch Swap & AdaIn & AdaIn+ & WCT   & Ours \\ \hline
		207.12s & 13.15s     & 0.49s & 0.16s  & 3.47s & 4.06s    \\ 
		\hline
	\end{tabular}
	\caption{Processing speed comparison.}
	\label{tab:perf}
\end{table}

\subsection{Qualitative Results}
\label{sec:quali}

\begin{figure*}
	\begin{center}
		\includegraphics[width=1.0\linewidth]{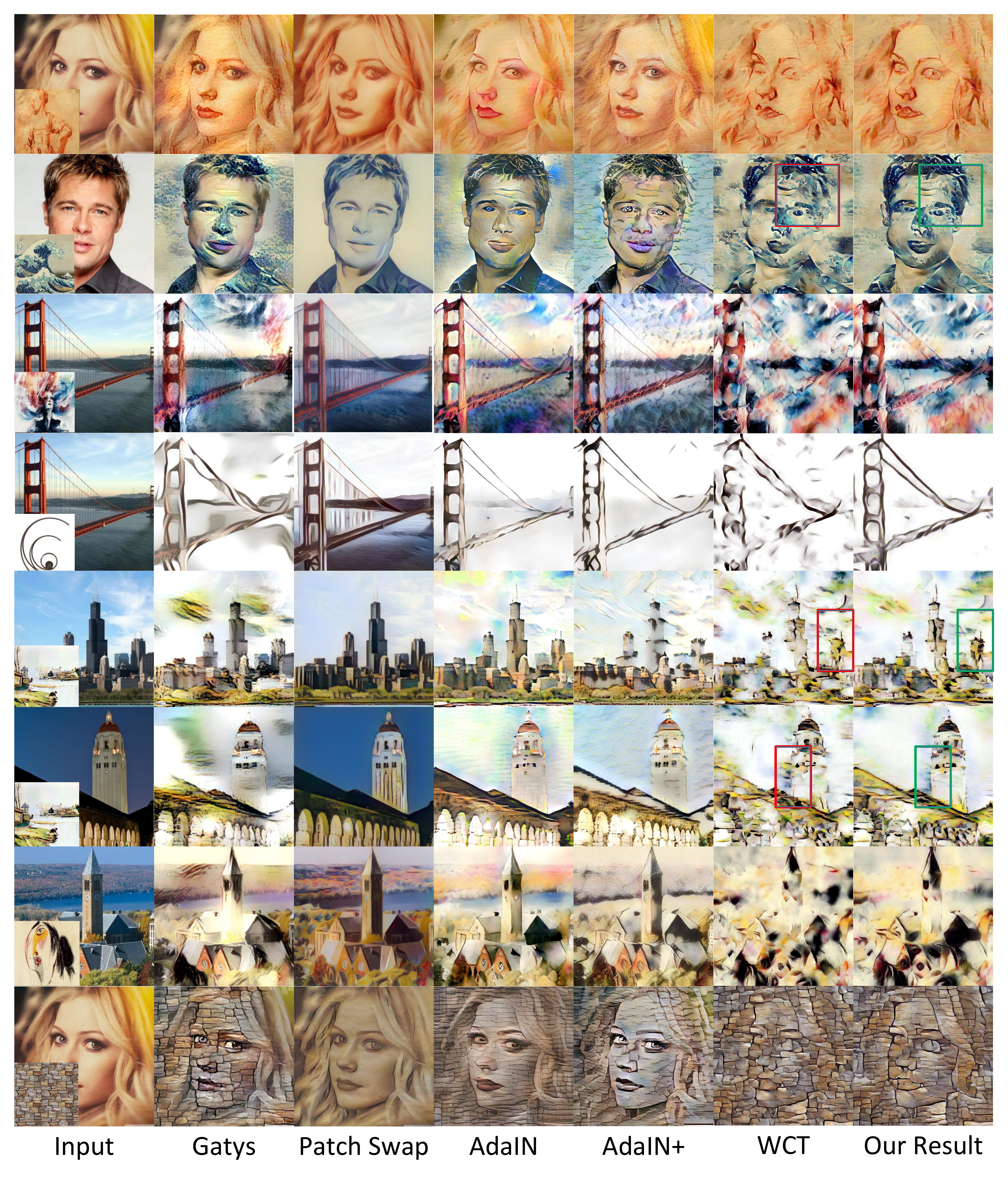}
	\end{center}
	\caption{Qualitative results. We compare our method against Gatys \cite{gatys2016image}, Patch Swap \cite{chen2016fast}, AdaIN (with our decoder) \cite{huang2017arbitrary}, AdaIN+ (with their decoder) \cite{huang2017arbitrary}, and WCT \cite{li2017universal}. AdaIN ignores the non-diagonal elements of covariance matrices, which results in less stylized output. WCT does not consider the content loss and cannot well-preserve the structure of content image as shown in the red rectangles. Our method can achieve both stylized and content-preserving results.}
	\label{fig:quali}
\end{figure*}

{\bf Our Method versus Gatys: } Gatys \cite{gatys2016image} is the pioneering work of neural style transfer and it can handle arbitrary styles. Although it uses time-consuming back-propagation to minimize the content loss and style loss, we still compare with it since its formulation is the foundation of our method. As shown in Figure \ref{fig:quali}, Gatys can usually achieve reasonable results, however, these results are not so stylized since the iterative solver cannot reach the optimal solution in limited iterations. Instead, our method tries to find the closed-form solution, which explicitly minimizes the style loss and content loss. Comparatively, our results are more stylized and they also well-preserve the structures of content images.  

{\bf Our Method versus Patch Swap: } As far as we know, Patch Swap \cite{chen2016fast} is the first work to use the encoder-feature transform-decoder framework. It chooses a certain layer of VGG network as the encoder and trains the corresponding decoder. The feature transform is formulated as neural patch swap. However, neural patch swap using the original feature tends to simply reconstruct the feature, thus the results are not stylized. Besides, Patch Swap only transfers the style in a certain layer, which also reduces the style transfer effect. \cite{sheng2018avatar} proposes to match the neural patch in the projected domains, for example, the whitened feature \cite{li2017universal}. Apart from this, \cite{sheng2018avatar} uses multiple layers to transfer the style, achieving more stylized results. Our work does not use the idea of neural patch match, instead, we focus on the theoretical analysis to deliver the closed-form solution. As can be seen in Figure \ref{fig:quali}, our result is more stylized compared with Patch Swap.

{\bf Our Method versus AdaIN and AdaIN+: } As discussed in the motivation, AdaIN \cite{huang2017arbitrary} assumes the channels of CNN feature are independent. For each channel, AdaIN matches two one-dimensional Gaussian distributions. However, the channels of CNN feature are actually correlated. Therefore, using AdaIN as the feature transform cannot achieve visually stylized results. Instead of using AdaIN as the feature transform method, AdaIN+ \cite{huang2017arbitrary} trains a decoder on large collections of content and style images. Although AdaIN+ only transfers the feature in a certain layer, it trains the decoder with style losses defined in multiple layers. We conduct the comparisons with both AdaIN and AdaIN+. As illustrated by Figure \ref{fig:quali}, the results of AdaIN and AdaIN+ are similar and both of them fail to achieve visually pleasing transfer results. Therefore, we believe the reason why AdaIN and AdaIN+ fail is because they ignore the correlation between channels of CNN feature. Instead, our work considers the correlation thus achieves more stylized results as shown in Figure \ref{fig:quali}.

\begin{figure}[t]
	\begin{center}
		\includegraphics[width=0.8\linewidth]{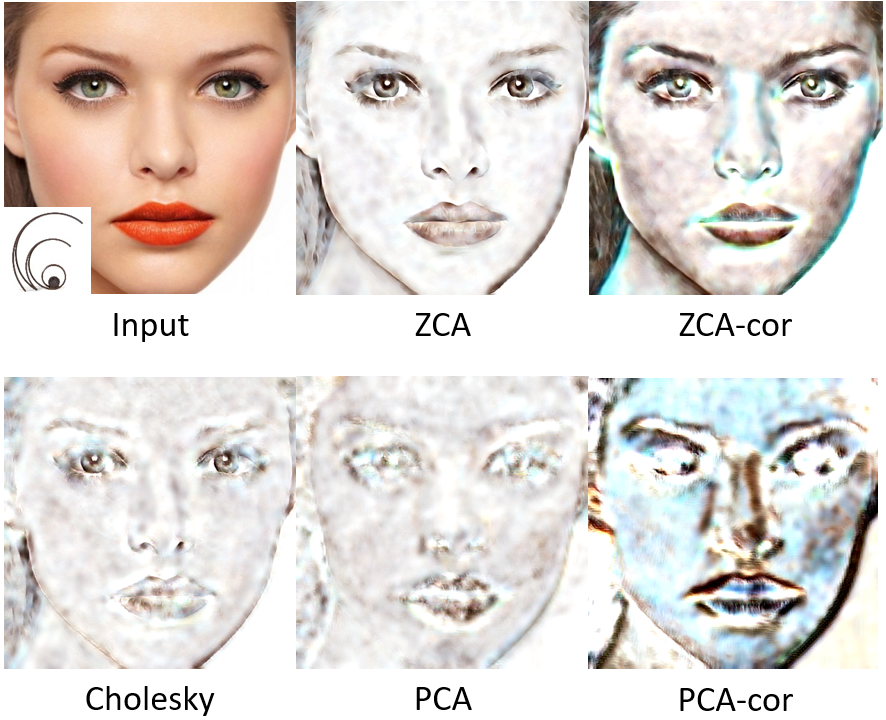}
	\end{center}
	\vspace{-3mm}
	\caption{Illustration of different whitening methods. We test some whitening methods with the feature of ReLU3\_1. As can be seen, ZCA whitening achieves better results. “cor” means correlated and details of whitening methods can be found in \cite{kessy2018optimal}}
	\label{fig:evalWhite}
\end{figure}

\begin{figure}[t]
	\begin{center}
		\includegraphics[width=0.8\linewidth]{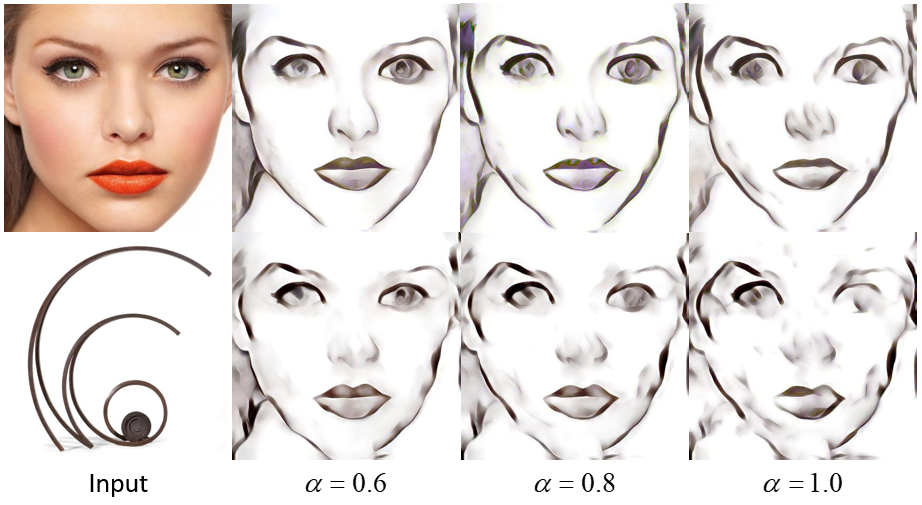}
	\end{center}
	\vspace{-3mm}
	\caption{Illustration of linear interpolation. The top row is the results of our method and the bottom row is the results of WCT. Linearly combining content feature with the transformed feature can help preserve the structure. With smaller $\alpha$ in Eq. \ref{eq:13}, WCT can preserve more structure. However, there is still obvious artifact. Instead, our method consistently achieves pleasing results.}
	\label{fig:evalWCT}
\end{figure} 

{\bf Our Method versus WCT: } WCT \cite{li2017universal} proposes to use feature whitening and coloring as the solution to style transfer. It chooses ZCA whitening in the paper and we test some other whitening methods with the feature of ReLU3\_1 as shown in Figure \ref{fig:evalWhite}. As can be seen, only ZCA whitening achieves reasonable results. This is because ZCA whitening is the optimal choice, which minimizes the difference between content feature and the whitened feature. Although the ZCA-whitened image can preserve the structure of content image, there is none constraint on the final transformed feature. Contrary to that, we consider to minimize the difference between content feature and the final transformed feature. As we have analyzed in the motivation section, WCT satisfies Eq. \ref{eq:2}. Therefore, it perfectly matches two high-dimension Gaussian distributions. However, it ignores the content loss of Gatys. Instead, we seek the closed-form solution, which additionally minimizes the content loss. As can be seen in Figure \ref{fig:quali}, our transformation can preserve better structures (see the red rectangles).

We also notice that the final feature can be the linear combination of original content feature and the transformed feature as shown in Eq. \ref{eq:13}. Where $\alpha$ is the weight of transformed feature. 

\begin{equation}
\begin{aligned}
{t^ * }(u) = \alpha t(u) + (1 - \alpha )u
\label{eq:13}
\end{aligned}
\end{equation}

\begin{table*}[t]
	\centering
	\begin{tabular}{|c|c|c|c|c|c|c|c|c|c|}
		\hline
		Method      &Gatys & Patch Swap & AdaIN & AdaIN+ & WCT*   &  Ours* \\ \hline
		Content Loss&0.096  & 0.086      & 0.167 & 0.151  & {\bf 0.296} &  {\bf 0.255}	\\ \hline
		Style Loss-1&23.77 & 100.8      & 15.85 & 15.9   & 3.89   &	3.60	\\ \hline
		Style Loss-2&8577.04& 30647.6    & 5351.6& 3355.5 & 594.4 &	457.8	\\ \hline
		Style Loss-3&6749.7& 15607.2    & 4564.7& 4905.5 & 1226.6 &	1203	\\ \hline
		Style Loss-4&325939& 562192     & 245133& 202767 & 187907 &	129695	\\ \hline
		Style Loss-5&15.96 & 17.73      & 14.1  & 12.48  & 24.33  &	12.37	\\
		\hline
	\end{tabular}
	\caption{Average content loss and style losses. * means fully matching the statistics of content and style features.}
	\label{tab:loss}
\end{table*}

\begin{figure*}
	\begin{center}
		\includegraphics[width=0.6\linewidth]{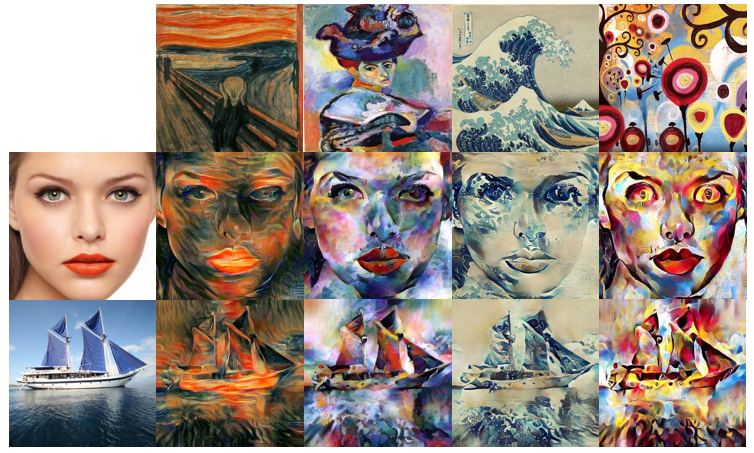}
	\end{center}
	\caption{More results. We show more results, where $\alpha$ is set as 1.}
	\label{fig:moreRes}
\end{figure*}

\begin{figure*}
	\begin{center}
		\includegraphics[width=0.6\linewidth]{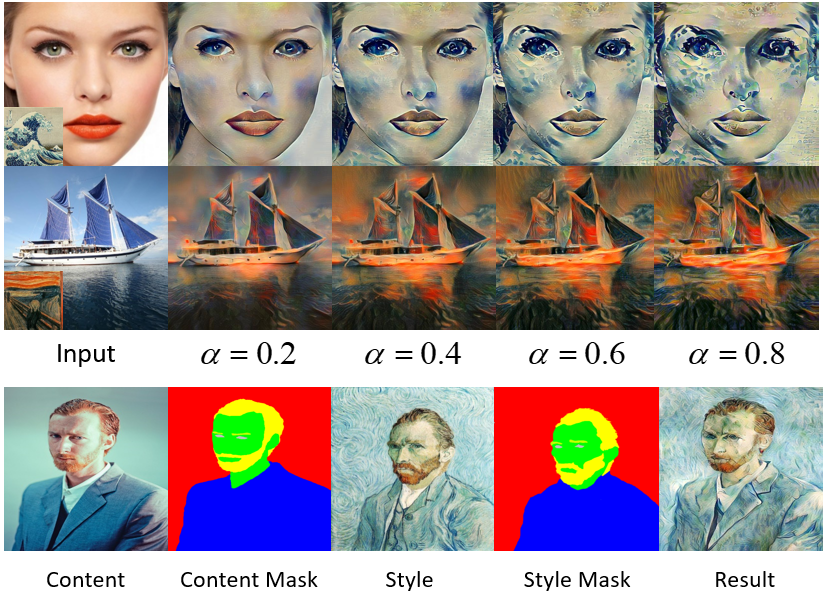}
	\end{center}
	\caption{Linear interpolation and semantic style transfer. Although we assume the neural features are sampled from MVG distributions in the proof, these results are all visually pleasing, which demonstrate the generalization ability of our work.}
	\label{fig:linearAndSem}
\end{figure*}

We show the results of different $\alpha$ values in Figure \ref{fig:evalWCT}. As illustrated by Figure \ref{fig:evalWCT}, adjusting the weight can change the degree of style transfer. With smaller $\alpha$, WCT can preserve more structure of content image. However, there is still obvious artifact even with small $\alpha$. Instead, our method consistently achieves visually pleasing results.  

\subsection{Quantitative Results}
\label{sec:quanti}

\begin{table}
	\begin{tabular}{|c|c|c|c|c|c|}
		\hline
		Gatys & Patch Swap & AdaIN & AdaIN+ & WCT  & Ours \\ \hline
		2.17  & 1.05       & 2.00  & 1.94   & 2.67 & 3.07 \\ 
		\hline
	\end{tabular}
	\caption{Average scores of user study.}
	\label{tab:user}
\end{table}

{\bf User Study: } Style transfer is a very subjective research topic. Although we have theoretically proved the advantages of our method, we further conduct a user study to quantitatively compare our work against Gatys, Patch Swap, AdaIN, AdaIN+ and WCT. This study uses 16 content images and 35 style images collected from published implementations, thus 560 stylized images are generated by each method. We show the content, style and stylized images to testers. We ask the testers to choose a score from 1 (worst) - 5 (best) for the purpose of evaluating the quality of style transfer. We do this user study with 50 testers online. The average scores are listed in the Table \ref{tab:user}. This study shows that our method improves the results of former works.


{\bf Content Loss and Style Loss: } In addition to user study, we also evaluate the content loss and style loss defined by Gatys \cite{gatys2016image}. We calculate the average content loss and style loss with the images of user study for each method. We normalize the content loss with the number of neural activations. The average losses are listed in Table \ref{tab:loss}. As can be seen, compared with WCT, our method achieves lower content loss and similar style loss. As for Gatys, Patch Swap, AdaIN and AdaIN+, they fail to achieve stylized results with high style losses as we have analyzed in the qualitative comparison part.

\subsection{More Results}
\label{sec:more}

We show more results to demonstrate the generalization of our method in Figure \ref{fig:moreRes}, where $\alpha$ is set as 1. To further evaluate the linear interpolation, we show two samples with different $\alpha$ values in Figure \ref{fig:linearAndSem}. We also combine our method with semantic style transfer as shown in Figure \ref{fig:linearAndSem}. Although we assume the neural features are sampled from MVG distributions in the proof, these results are all visually pleasing, which demonstrate the generalization ability of the proposed method.

\subsection{Limitations}
\label{sec:limit}

Our method still has some limitations. For example, we evaluate the frame-by-frame results of video style transfer. Although our method can preserve better structure compared with former works, the frame-by-frame results still contain obvious jittering. We find that the temporal jittering is not only caused by feature transform but also caused by the information loss of encoder networks. Deep encoder network will cause obvious temporal jittering even without feature transform.

Besides, style transfer is a very subjective problem. Although the Gram matrix representation proposed by Gatys has been widely used, mathematically modeling of what people really feel about style is still an unsolved problem. Exploring the relation between deep neural network and image style is an interesting topic.

\section{Conclusion}

In this paper, we first present a novel interpretation of neural style transfer by treating it as an optimal transport problem. Then we demonstrate the theoretical relations between our interpretation and former works, for example, AdaIN and WCT. Based on our formulation, we derive the unique closed-form solution by additionally considering the content loss. Our solution preserves better structure compared with former works due to the minimization of content loss. We hope this paper can inspire future works in style transfer.

{\bf Acknowledgements.} This work was supported by the National Key R\&D Program of China 2018YFA0704000, the NSFC (Grant No. 61822111, 61727808, 61671268, 61132007, 61172125, 61601021, and U1533132) and Beijing Natural Science Foundation (L182052).

{\small
\bibliographystyle{ieee_fullname}
\bibliography{egbib}
}

\end{document}